\documentclass[a4paper,twocolumn,fleqn]{article}

\usepackage{ist}
\usepackage{graphicx}
\usepackage{amsmath}
\usepackage{amssymb}

\usepackage{multirow}
\usepackage{rotating}
\usepackage{booktabs}
\usepackage{colortbl}
\usepackage[table]{xcolor}

\usepackage{listings}
\usepackage[most]{tcolorbox}
\usepackage{enumitem}
\usepackage{url}
\usepackage{subcaption}
\usepackage{subcaption}
\usepackage{caption}
\usepackage{placeins}
\usepackage{tcolorbox}
\usepackage{enumitem}

\setlist[itemize]{leftmargin=*, itemsep=1pt, topsep=2pt, parsep=0pt, partopsep=0pt}
\setlist[enumerate]{leftmargin=*, itemsep=1pt, topsep=2pt, parsep=0pt, partopsep=0pt}

\title{Evaluating Vision-Language Models for Image Quality Assessment using Psychophysical Data}
 
\author{
Imran Mehmood\textsuperscript{1,2},
Imad Ali Shah\textsuperscript{1,2},
Ming Ronnier Luo\textsuperscript{3},
and Brian Deegan\textsuperscript{1,2}\\
\small \textsuperscript{1} School of Engineering, University of Galway, Ireland\\
\small \textsuperscript{2} Lero the Research Ireland Centre for Software, Ireland\\
\small \textsuperscript{3} State Key Laboratory of Extreme Photonics and Instrumentation, Zhejiang University, Hangzhou, China.
}

\begin{document}

\maketitle

\begin{abstract}
Psychophysical experiments remain the most reliable approach for perceptual image quality assessment (IQA), yet their cost and limited scalability motivate automated alternatives. This paper investigates whether Vision-Language Models (VLMs) can assist in assessing perceived image appearance and quality. We introduce a psychophysics-inspired framework to probe VLM perceptual sensitivity through controlled pairwise image comparisons of contrast, colorfulness, and overall preference. Six VLMs (four proprietary and two open-weight models) are compared against psychophysical data. Results reveal strong attribute-dependent variability: Claude exhibits the highest internal consistency, whereas GPT achieves the strongest agreement for overall preference. Claude and Qwen show the strongest alignment for colorfulness, while Qwen performs best for contrast. However, no model consistently matches human perception across all attributes. High self-consistency does not necessarily imply perceptual validity, and VLM--human agreement generally improves when perceptual differences among renderings are more pronounced.
\end{abstract}

\section{Introduction}
\label{sec:intro}

Evaluation of perceptual image quality (IQ)~\cite{ghadiyaram2017perceptual,mehmood2022reference} in computational imaging remains a central challenge for researchers. Numerous IQ evaluation metrics~\cite{fang2014no,ghadiyaram2017perceptual,mittal2012no,xue2014blind,xu2016blind,khan2026no} have been developed depending on the research area to model human perception and quantify IQ. The correlation of these metrics is highly scene-content dependent, and the metrics struggle to correlate well with human perception for complex scenes~\cite{khan2026no}. Consequently, psychophysical evaluation with human observers remains the most reliable method, despite being time-consuming, expensive, and difficult to scale.

Traditionally, IQA methods include objective and subjective approaches. Objective methods range from conventional metrics such as PSNR and SSIM to perceptually motivated and learned approaches~\cite{chow2016review,preiss2014color,zhang2018unreasonable}, although their predictions may not fully reflect perceived image quality.

Recent advances in vision-language models (VLMs) have significantly improved machine understanding of visual content. Beyond recognition and captioning, these models can generate qualitative assessments of image appearance attributes, raising the question of whether they can approximate human perceptual evaluations. If reliable, VLM-based assessments could accelerate algorithm development by enabling rapid screening and hypothesis testing before costly human studies. However, perceptual IQ assessment (IQA) presents unique challenges. Assessments of image appearance attributes such as contrast and colorfulness are subjective, context-dependent, and shaped by subtle interactions, necessitating a controlled experimental environment.

In the context of image quality, recent benchmarks have also begun to examine whether foundation models can assess low-level and perceptual visual attributes. Studies such as Q-Bench~\cite{wu2024qbench} evaluate the ability of foundation models to judge low-level vision and image quality, while related work on multi-image and comparative reasoning~\cite{kil2024mllm,kim2026vlmsubtlebench} shows growing interest in whether VLMs can detect subtle visual differences. However, perceptual IQ assessment presents unique challenges: attributes such as contrast and colorfulness are subjective, context-dependent, and influenced by subtle visual interactions. Leveraging VLMs in this domain, particularly to evaluate their ability to capture perceptual cues and identify failure conditions, therefore remains an important and underexplored direction.

In this work, four proprietary and two open-weight VLMs are evaluated by directly comparing their perceptual assessments with psychophysical data using the same pairwise forced-choice framework. A psychophysics-inspired approach is adopted to probe the perceptual sensitivities of VLMs under controlled conditions. Three perceptual scales, i.e., contrast, colorfulness, and overall preference, are considered to perform a diagnostic analysis to determine how reliably these attributes are captured by VLMs and to identify where their judgments diverge from those of human observers. Unlike conventional IQA benchmarks that primarily evaluate prediction accuracy, this study examines whether VLM judgments reproduce human perceptual behavior under the same psychophysical comparison framework. Our evaluation examines model behavior along three axes: internal consistency, agreement across models, and alignment with human perceptual rankings.

\section{Methodology}

\subsection{Psychophysical Dataset}
\begin{figure}[!t]
  \centering
  \includegraphics[width=\columnwidth]{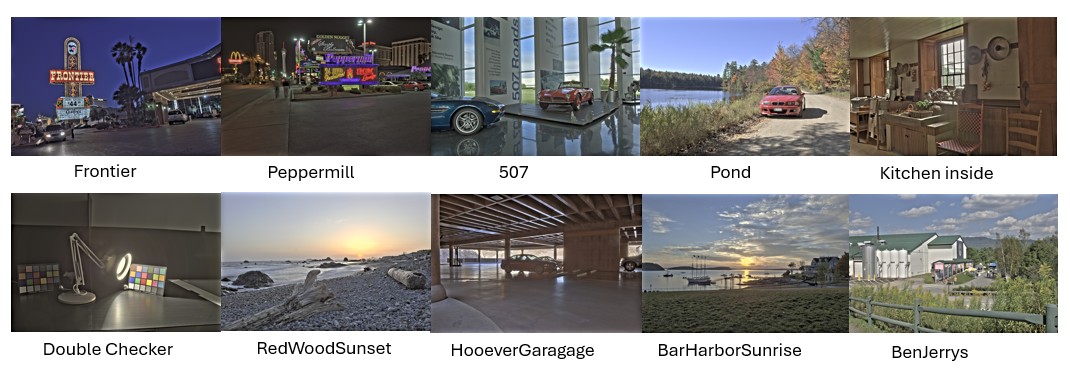}
  \caption{Experiment Images.}
  \label{fig:experimentImages}
\end{figure}

The dataset for this study is based on a psychophysical experiment conducted by Mehmood et al.~\cite{mehmood2023perceptual} on the IQ evaluation of tone mapping operators (TMOs) in terms of three perceptual IQ scales: contrast, colorfulness, and overall preference. The evaluation dataset comprised 10 HDR images rendered using seven TMOs with different tone-reproduction and color-processing characteristics, as detailed in~\cite{mehmood2024generic}: Hui~\cite{hui2018}, Khan~\cite{khan2020}, Liang~\cite{liang2018}, Meylan~\cite{meylan2006}, Reinhard~\cite{reinhard2002}, Schlick~\cite{schlick1995}, and $\mathrm{TMO}_{z}$~\cite{mehmood2023perceptual}, representing diverse scene content, illumination conditions, and dynamic-range characteristics, as shown in Figure~\ref{fig:experimentImages}. These models were selected to provide a stimulus space where the rendered images for each scene exhibit both subtle and significant variations across the three target IQ scales.  Each HDR image was rendered using default parameters for each TMO, resulting in a total of 70 images (10 HDR images × 7 TMOs). The TMOs were not used for ranking purposes, but to provide a controlled stimulus space and existing psychophysical reference data for evaluating VLMs; the relevant stimulus preparation and evaluation settings are reported below to support reproducibility.

\subsection{Psychophysical Experiment}
The experimental settings and assessment protocols were as follows:

\begin{itemize}
\item \textbf{Observers and viewing conditions:} The experiment involved 20 observers (mean age 25, SD = 2.5; 7 female, 13 male) with normal or corrected-to-normal vision, all passing the Ishihara color vision test~\cite{ishihara1985ishihara}. Experiments were conducted in a dark-room environment using an Apple Pro Display XDR calibrated to 562 cd/m$^{2}$ (CIE D65, 1931 observer). Observers followed ITU viewing conditions~\cite{itu2020}, with a 1.2 m viewing distance ($\approx$2.5$\times$ display height) and 22$^\circ$ viewing angle.

\item \textbf{Display calibration and stimulus preparation:} Display uniformity was verified using a 3×3 grid with a mean $\Delta E_{ab}$ of 1.1. Gain-offset-gamma (GOG) model~\cite{berns1996} was developed to characterize the display using a 24-patch ColorChecker~\cite{mccamy1976color}, achieving a mean characterization error of 0.39 $\Delta E_{ab}$ (range: 0.16–1.13). 

\item \textbf{Experimental protocol and data processing:} The images were transformed into the display RGB space prior to presentation. Observers performed pairwise comparisons on three perceptual scales: contrast, colorfulness, and overall preference. Each observer evaluated 273 image pairs (2,730 total assessments including repetitions). Data reliability was confirmed via intra- and inter-observer variability analysis. Responses were converted to Z-scores using Thurstone’s law of comparative judgment~\cite{thurstone1927} to derive perceptual rankings of the images for each of the three perceptual scales.
\end{itemize}

\subsection{VLMs Assessment}
A set of six recent VLMs is evaluated, including both proprietary and open-weight models: Anthropic Claude Opus 4.6 (Claude)~\cite{Anthropic_ClaudeOpus46_2026}, Google Gemini 3.1 Pro (Gemini)~\cite{Google_Gemini31Pro_2026}, OpenAI GPT-5.2 (GPT)~\cite{OpenAI_GPT52_2026}, xAI Grok-4.1 (Grok)~\cite{xAI_Grok41_2026}, OpenGVLab InternVL-3.5-38B (Intern)~\cite{OpenGVLab_InternVL35_38B_2026}, and Alibaba Qwen3-VL-32B-Instruct (Qwen)~\cite{QwenVL3_2026}.

\subsubsection{Prompt Design}
To design effective and consistent prompts across all VLMs, explicit definitions of contrast, colorfulness, and overall image quality were first elicited to assess whether these models interpret such perceptual attributes in a manner consistent with human observers and established CIE definitions. This preliminary step ensured that model responses were grounded in well-defined and comparable perceptual criteria.

The robustness of VLM-based assessments was further evaluated under prompt rephrasing and repeated sampling, while maintaining identical underlying evaluation queries. To systematically investigate this behavior, a structured prompt framework was formulated to elicit comparative quality judgments. The framework employs a forced-choice paradigm aligned with the paired comparison protocol commonly used in psychophysical experiments. The complete prompt template is provided below:

\subsubsection{Assessment Protocol}

Tone-mapped image pairs were submitted to VLMs in their original 8-bit sRGB encoding using each API's native multi-

\begin{tcolorbox}[
    colback=gray!5, 
    colframe=gray!80!black, 
    fonttitle=\bfseries,
    title=Prompt,
    arc=0mm,                
    boxrule=0.4pt,          
    left=1mm,               
    right=0.75mm
]
\footnotesize
Compare these two images:
\begin{itemize}[leftmargin=*, noitemsep, topsep=2pt]
    \item The first image is `A'
    \item The second image is `B'
\end{itemize}
\vspace{0.25mm}
For each of the following IQ scales, select exactly one image and provide a brief rationale:
\begin{itemize}[leftmargin=*, noitemsep, topsep=2pt]
    \item Which image has higher Contrast (choose A or B)
    \item Which image has higher Colorfulness (choose A or B)
    \item Which image has better Overall image quality (choose A or B)
\end{itemize}
\vspace{0.25mm}
Each key \textbf{MUST} contain a nested object with two keys: `selection' (value should be strictly A or B) and `reason' (a string).
\end{tcolorbox}

\noindent image input, without additional preprocessing or color conversion. The image names were specifically changed to alphabets, i.e., A-G, to remove any bias arising from the original image source or if the VLMs had learned those 
rendering models. Hence, each comparison was conducted anonymously and independently without reference to the source or previous comparisons, maintaining experimental independence.
Each pairwise comparison was repeated three times using default model settings to quantify variability across repeated assessments. API calls were performed using the default configuration, with the temperature set to 1, which is the API default, to maintain consistent experimental settings while permitting natural variation across repeated model responses.

Overall, the experiment resulted in
$\binom{7}{2}\times10\times6\times3=3780$ assessments. For each assessment, the selected image pair (e.g., A or B) and the textual rationale for each attribute were extracted and processed.

\section{Analysis and Results}

To obtain rankings for each image quality scale, VLM preference scores are aggregated across all pairwise comparisons and converted into standardized \(Z\)-scores using Thurstone’s law of comparative judgment~\cite{thurstone1927}, following an approach similar to the psychophysical evaluation methodology in Mehmood et al.~\cite{mehmood2023perceptual}.

\begin{table*}[!t]
\centering
\caption{Intra-model variability performance comparison (VR~\%): Mean (Max, SD).}
\label{tab:model-performance-consolidated}
\vspace{0pt}
\small
\resizebox{\textwidth}{!}{%
\begin{tabular}{lcccccc}
\toprule
\textbf{Scale} & \textbf{Claude} & \textbf{Gemini} & \textbf{GPT} & \textbf{Grok} & \textbf{Intern} & \textbf{Qwen} \\
\midrule
\textbf{Contrast} & 1.90 (2.38, 0.82) & 2.54 (2.86, 0.55) & 4.44 (5.24, 0.73) & 6.67 (7.62, 0.95) & 17.78 (20.00, 1.98) & 2.54 (2.86, 0.27) \\
\textbf{Colorfulness} & 0.95 (1.43, 0.48) & 6.03 (7.14, 1.53) & 8.89 (10.00, 0.99) & 7.30 (8.10, 1.37) & 14.92 (17.14, 2.40) & 2.86 (4.29, 1.26) \\
\textbf{Overall} & 1.27 (1.43, 0.27) & 10.48 (13.81, 4.97) & 7.62 (8.10, 0.82) & 8.25 (10.48, 2.15) & 13.02 (14.29, 1.45) & 6.35 (7.62, 1.45) \\
\bottomrule
\end{tabular}
}
\vspace{20 pt}   
\end{table*}

\begin{table*}[t]
\centering
\caption{Cross-model variability rate comparison (VR~\%): Mean (Max, SD)}
\label{tab:Inter-VLM-Consistency}
\small
\resizebox{0.98\textwidth}{!}{
\begin{tabular}{llccccc}
\toprule
\textbf{Scale} & \textbf{Model} & \textbf{Claude} & \textbf{GPT} & \textbf{Gemini} & \textbf{Grok} & \textbf{Intern} \\
\midrule

\multirow[t]{5}{*}{\textbf{Contrast}}
& \textbf{GPT} & $25.87\ (26.67,\ 0.67)$ & --- & --- & --- & --- \\
& \textbf{Gemini} & $25.71\ (26.67,\ 0.79)$ & $10.00\ (11.90,\ 1.56)$ & --- & --- & --- \\
& \textbf{Grok} & $46.03\ (46.67,\ 0.53)$ & $43.02\ (43.81,\ 0.98)$ & $41.38\ (42.86,\ 1.08)$ & --- & --- \\
\arrayrulecolor{gray}
\cline{2-7}
\arrayrulecolor{black}
& \textbf{Intern} & $40.79\ (40.95,\ 0.24)$ & $52.06\ (52.38,\ 0.24)$ & $48.73\ (49.05,\ 0.48)$ & $52.22\ (52.86,\ 0.48)$ & --- \\
& \textbf{Qwen} & $26.35\ (26.67,\ 0.24)$ & $24.92\ (25.24,\ 0.24)$ & $21.90\ (22.86,\ 0.82)$ & $38.41\ (39.05,\ 0.48)$ & $49.05\ (49.05,\ 0.00)$ \\
\midrule

\multirow[t]{5}{*}{\textbf{Colorfulness}}
& \textbf{GPT} & $21.90\ (23.33,\ 1.17)$ & --- & --- & --- & --- \\
& \textbf{Gemini} & $38.73\ (40.95,\ 1.39)$ & $28.99\ (33.33,\ 2.31)$ & --- & --- & --- \\
& \textbf{Grok} & $38.94\ (40.48,\ 0.95)$ & $36.19\ (38.10,\ 1.26)$ & $37.57\ (39.05,\ 1.08)$ & --- & --- \\
\arrayrulecolor{gray}
\cline{2-7}
\arrayrulecolor{black}
& \textbf{Intern} & $21.59\ (21.90,\ 0.24)$ & $32.70\ (34.76,\ 1.56)$ & $51.43\ (53.81,\ 2.06)$ & $44.44\ (45.24,\ 1.19)$ & --- \\
& \textbf{Qwen} & $15.08\ (15.71,\ 0.63)$ & $25.24\ (27.14,\ 1.89)$ & $41.75\ (42.86,\ 0.86)$ & $32.22\ (32.86,\ 0.63)$ & $21.43\ (21.43,\ 0.00)$ \\
\midrule

\multirow[t]{5}{*}{\textbf{Overall}}
& \textbf{GPT} & $35.08\ (36.67,\ 1.06)$ & --- & --- & --- & --- \\
& \textbf{Gemini} & $39.21\ (45.24,\ 4.31)$ & $20.21\ (22.86,\ 1.83)$ & --- & --- & --- \\
& \textbf{Grok} & $49.21\ (50.00,\ 0.48)$ & $34.13\ (36.67,\ 1.61)$ & $34.44\ (37.14,\ 2.69)$ & --- & --- \\
\arrayrulecolor{gray}
\cline{2-7}
\arrayrulecolor{black}

& \textbf{Intern} & $24.60\ (25.24,\ 0.63)$ & $53.33\ (54.29,\ 0.71)$ & $54.60\ (61.90,\ 5.54)$ & $65.24\ (66.19,\ 1.09)$ & --- \\
& \textbf{Qwen} & $19.21\ (19.52,\ 0.24)$ & $37.78\ (39.52,\ 1.67)$ & $40.95\ (42.86,\ 1.49)$ & $44.29\ (45.71,\ 1.24)$ & $24.76\ (24.76,\ 0.00)$ \\

\bottomrule
\end{tabular}
}
\end{table*}

\subsection{Intra-Model Variability Analysis}
To assess VLM reliability, intra-model consistency was quantified across three repeated assessments for each image pair. For each VLM and perceptual scale, the inter-run variability rate (VR\%) was defined as the percentage of trials that produced different rankings across the three runs. Accordingly, lower VR indicates more repeatable model judgments, whereas higher VR indicates greater variation across repeated assessments, analogous to intra-observer variability in psychophysical experiments.

Table~\ref{tab:model-performance-consolidated} summarizes intra-model VR across the three perceptual scales. Claude demonstrates the highest consistency, with very low variability across the three perceptual scales, i.e., Colorfulness (0.95\%), Overall Preference (1.27\%), and Contrast (1.90\%), indicating highly repeatable perceptual judgments. 

Gemini shows increased variability when the task shifts to holistic judgments. While Contrast remains relatively stable (2.54\%), variability rises for Colorfulness (6.03\%) and Overall Preference (10.48\%). GPT and Grok exhibit moderate variability across scales, suggesting reasonably stable but less consistent perceptual evaluations.

In contrast, Intern displays the highest variability across all perceptual dimensions, indicating greater stochasticity in its decision process. Qwen maintains low variability for Contrast and Colorfulness but shows higher variability for Overall Preference, suggesting that holistic judgments remain more challenging.

Most VLMs demonstrate strong internal consistency, with variability typically below 10\%. The observed variability tends to arise in perceptually ambiguous cases where stimulus differences are subtle, reflecting patterns similar to decisional uncertainty observed in human psychophysical experiments.

\subsection{Cross-Model Variability Analysis}

To quantify agreement among VLMs independent of human judgments, pairwise variability rates were computed between all model pairs. For each perceptual scale, the percentage of image pairs for which two models selected different images as superior was calculated. This analysis reveals cross-model agreement patterns across the IQ scales to identify attribute-dependent differences in inter-model alignment. The results are summarized in Table~\ref{tab:Inter-VLM-Consistency}, which reports the mean VR percentage for each model along with the maximum and SD.

Consistent with the inter-observer variability reported in the psychophysical experiment~\cite{mehmood2023perceptual}, Contrast exhibits higher inter-model inconsistency than Colorfulness. For example, Grok shows substantial disagreement with Claude (46.03\%) and GPT (43.02\%), while the open-weight model Intern exhibits the highest disagreement with the other VLMs. Qwen has better agreement with proprietary models. These results suggest that contrast evaluation remains particularly challenging for cross-model alignment.

Analysis of individual model pairs reveals clusters of perceptual similarity and divergence. GPT and Gemini demonstrate relatively strong agreement for contrast (10\% disagreement), indicating a shared strategy for evaluating luminance-driven attributes. In contrast, Grok and Intern diverge more strongly from the proprietary model cluster, suggesting different computational sensitivities to tonal structure and local contrast.

\begin{table*}[t]
\centering
\caption{Summary of the qualitative rationale criteria used by the six evaluated VLMs.}
\label{tab:vlm_rationale_criteria}
\renewcommand{\arraystretch}{1.25}
\begin{tabular}{@{}p{0.13\linewidth} p{0.27\linewidth} p{0.25\linewidth} p{0.27\linewidth}@{}}
\toprule
\textbf{Scale} & \textbf{Primary Criteria (Rewarded)} & \textbf{Secondary Criteria} & \textbf{Penalty Cues (Downgraded)} \\
\midrule
Colorfulness &
Color saturation; vividness and vibrancy of hues &
Tonal richness; exposure balance; natural color rendering &
Muted, dull, washed-out, or desaturated colors; hazy/milky film; flat appearance \\
\addlinespace
Contrast &
Highlight--shadow separation; dynamic/tonal range &
Deep blacks and bright highlights; exposure balance; depth of shadows &
Flat or low-contrast appearance; washed-out tones; hazy/milky veil; lifted black levels \\
\addlinespace
Overall Quality &
Sharpness and detail; exposure balance; depth and clarity; naturalness &
Tonal range; highlight/shadow detail preservation; color fidelity &
Haze/milky film; washed-out or flat look; over-processed/HDR-like artifacts; loss of fine detail \\
\bottomrule
\end{tabular}
\end{table*}
\begin{figure*}[!t]
\centering
\includegraphics[
    width=\textwidth,
    height=0.40\textheight,
    keepaspectratio
]{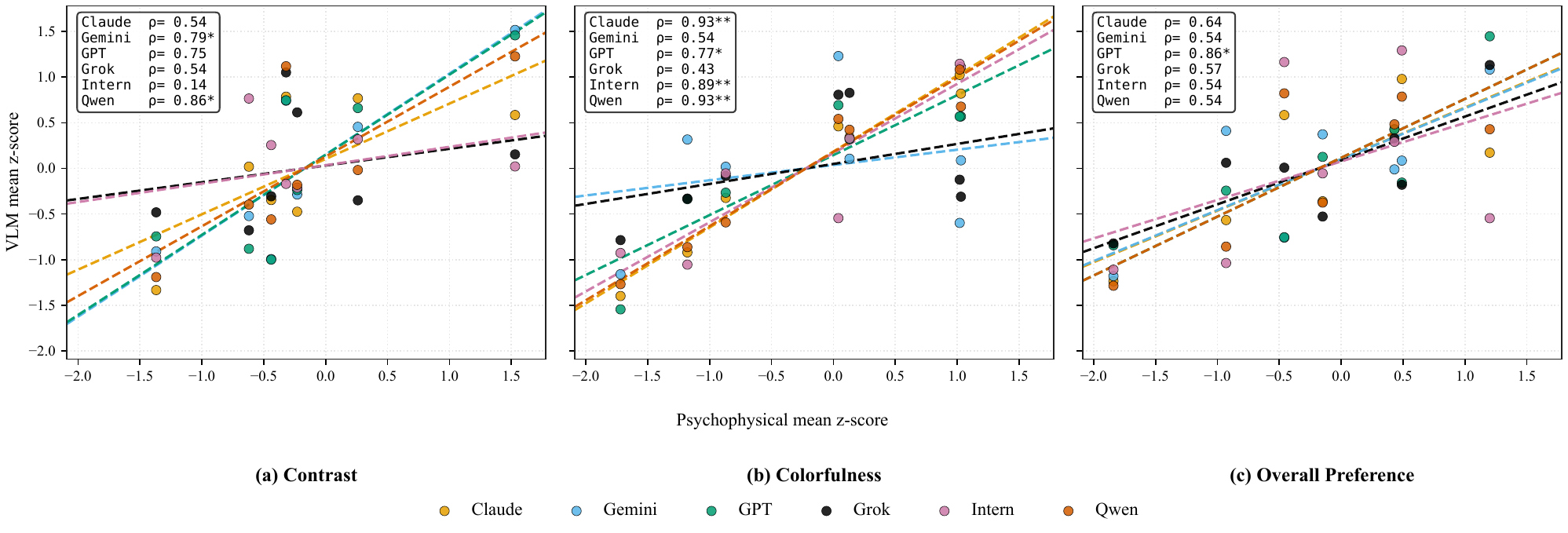}
\caption{Correlation between VLM and human mean rankings across the Contrast, Colorfulness, and Overall Preference scales.}
\label{fig:correlationFigures}
\end{figure*}

Colorfulness evaluations show slightly stronger agreement for certain model pairs. Claude and Qwen demonstrate the highest alignment (15.08\% disagreement), whereas Gemini and Intern display substantial divergence (51.43\%). In particular, Grok consistently shows higher disagreement relative to the Claude–GPT–Gemini cluster, approaching 50\% disagreement with Claude.

For the Overall Preference scale, inter-model variability increases as models integrate multiple perceptual cues into a single judgment. GPT and Gemini show relatively stronger agreement, suggesting similar strategies for holistic quality evaluation. Among the open-weight models, Qwen demonstrates comparatively better alignment with proprietary models, indicating closer agreement in overall preference decisions.

\subsection{Rationale Analysis}

Table~\ref{tab:vlm_rationale_criteria} summarizes the dominant perceptual factors extracted from the brief rationales explicitly requested from each VLM for every pairwise assessment. Across models, the rationales indicate a clear attribute-specific structure. For colorfulness, models primarily reward saturated, vivid, and vibrant hue reproduction, while secondary considerations include tonal richness, balanced exposure, and natural color rendering. Images described as muted, dull, washed out, desaturated, or affected by a hazy veil are consistently downgraded. 

For contrast assessments, the dominant criterion is the separation between highlights and shadows, often framed in terms of dynamic range or tonal range. Additional positive cues include deep blacks, bright highlights, balanced exposure, and shadow depth, whereas flat appearance, lifted black levels, washed-out tones, and haze are treated as negative indicators. 

For overall quality, the rationales are more integrative: VLMs emphasize sharpness, fine detail, exposure balance, clarity, depth, and naturalness, while also considering tonal range, highlight/shadow detail preservation, and color fidelity. Penalized cues include haze, flat or washed-out rendering, over-processed or HDR-like artifacts, and loss of fine detail. Overall, the thematic analysis suggests that VLMs rely on perceptually meaningful descriptors that broadly correspond to established image-quality attributes.

\subsection{Correlations with Psychophysical Data}
Figure~\ref{fig:correlationFigures} depicts scatter plots between VLM assessments and psychophysical data for each scale, along with correlations and p-values.

For the Contrast scale, the degree of concordance between VLMs and human psychophysical responses is heterogeneous across models. Qwen ($\rho = 0.86$, $p = 0.014$) and Gemini ($\rho = 0.79$, $p = 0.036$) exhibit statistically significant monotonic agreement with the human-derived contrast scale, indicating that these models preserve the ordinal structure of perceived contrast across the rendered images. GPT also demonstrates a relatively strong association ($\rho = 0.75$), although it does not reach conventional statistical significance ($p = 0.052$, $n = 7$). In contrast, Claude and Grok yield moderate but non-significant correlations ($\rho = 0.54$, $p = 0.215$), while Intern shows negligible agreement ($\rho = 0.14$, $p = 0.760$), suggesting an absence of a human-aligned contrast representation. 

The Colorfulness scale exhibits the highest level of agreement, with several VLMs achieving strong rank correspondence with psychophysical judgments. Claude and Qwen attain the highest correlations ($\rho = 0.93$, $p = 0.003$), closely followed by Intern ($\rho = 0.89$, $p = 0.007$), all of which indicate statistically robust alignment with the human colorfulness scale. GPT also demonstrates significant agreement ($\rho = 0.77$, $p = 0.041$), albeit with increased dispersion among intermediate rendered images. These results suggest that these models effectively encode chromatic attributes that are predictive of perceived colorfulness, likely reflecting sensitivity to low-level color statistics such as saturation and chroma distribution. In contrast, Grok ($\rho = 0.43$, $p = 0.337$) and Gemini ($\rho = 0.11$, $p = 0.819$) show moderate and weak positive correlations, respectively, neither of which is statistically significant, indicating poor chromatic perceptual alignment.

\begin{figure}[!h]
  \includegraphics[width=\columnwidth]{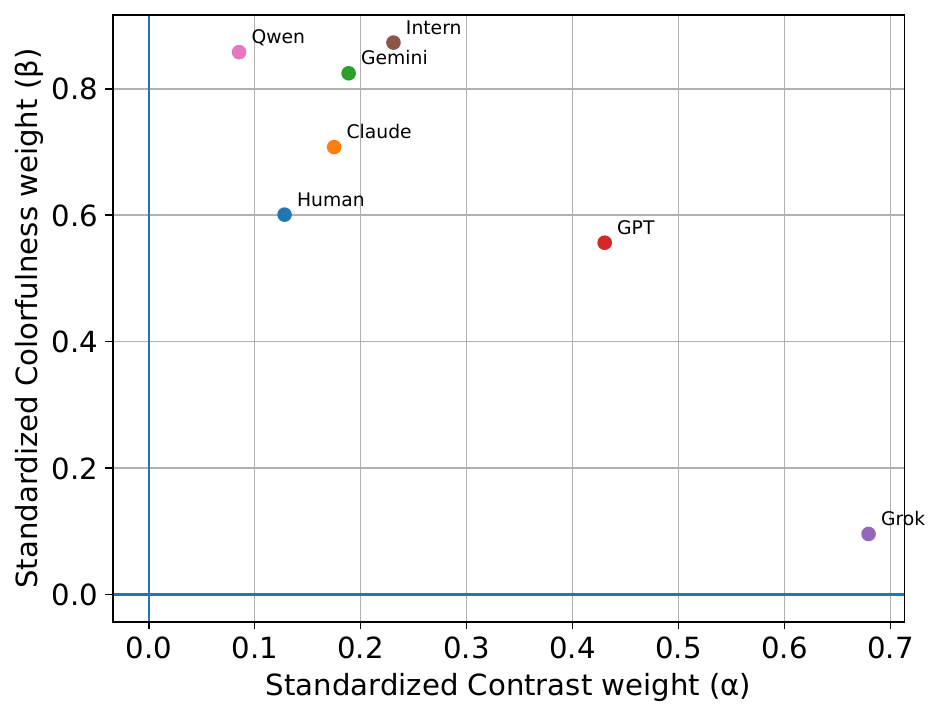}
  \caption{Attribute weighting for overall preference. The x-axis represents the contrast weight ($\alpha$) and the y-axis represents the colorfulness weight ($\beta$).}
  \label{fig:attribute_weights}
\end{figure}

For the Overall Preference scale, all VLMs exhibit positive monotonic associations with human psychophysical rankings, with Spearman coefficients ranging from $\rho = 0.54$ to $\rho = 0.86$. GPT achieves the highest agreement ($\rho = 0.86$, $p = 0.014$), indicating statistically significant alignment with perceptual preference ordering. The remaining models (Claude, Gemini, Grok, Intern, and Qwen) yield moderate correlations ($\rho \approx 0.54$--$0.64$) that do not reach statistical significance ($p > 0.05$), suggesting that these associations should be interpreted cautiously.

\begin{figure*}[!t]
\centering

\begin{minipage}[t]{0.32\textwidth}
    \centering
    \includegraphics[width=\linewidth]{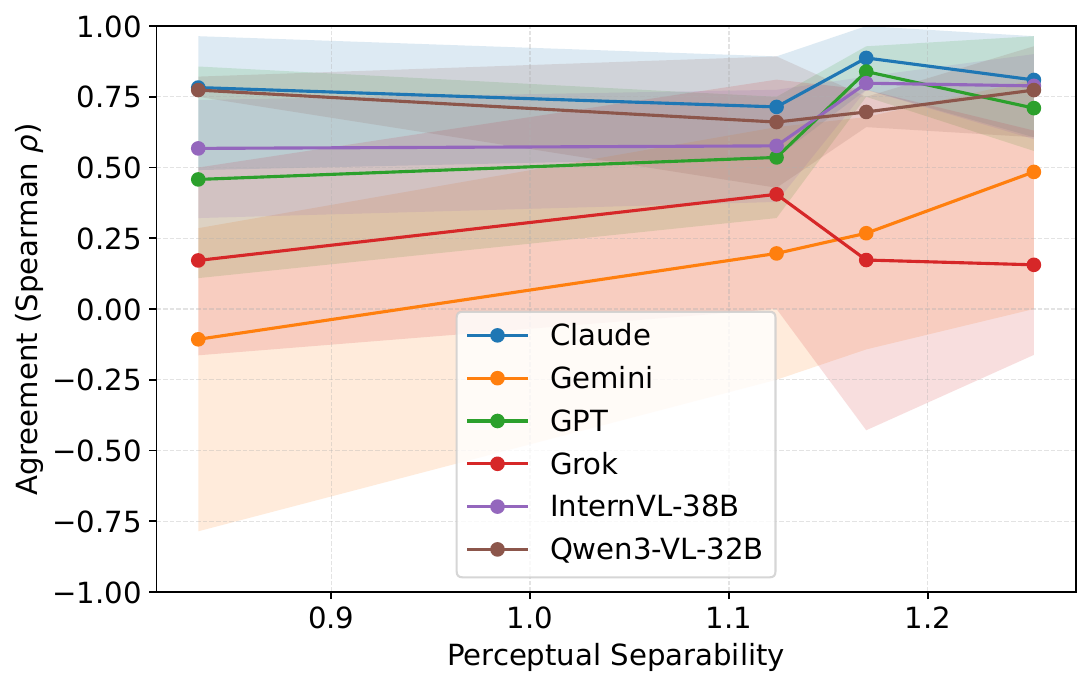}
    \caption*{(a) Colorfulness}
\end{minipage}%
\hfill
\begin{minipage}[t]{0.32\textwidth}
    \centering
    \includegraphics[width=\linewidth]{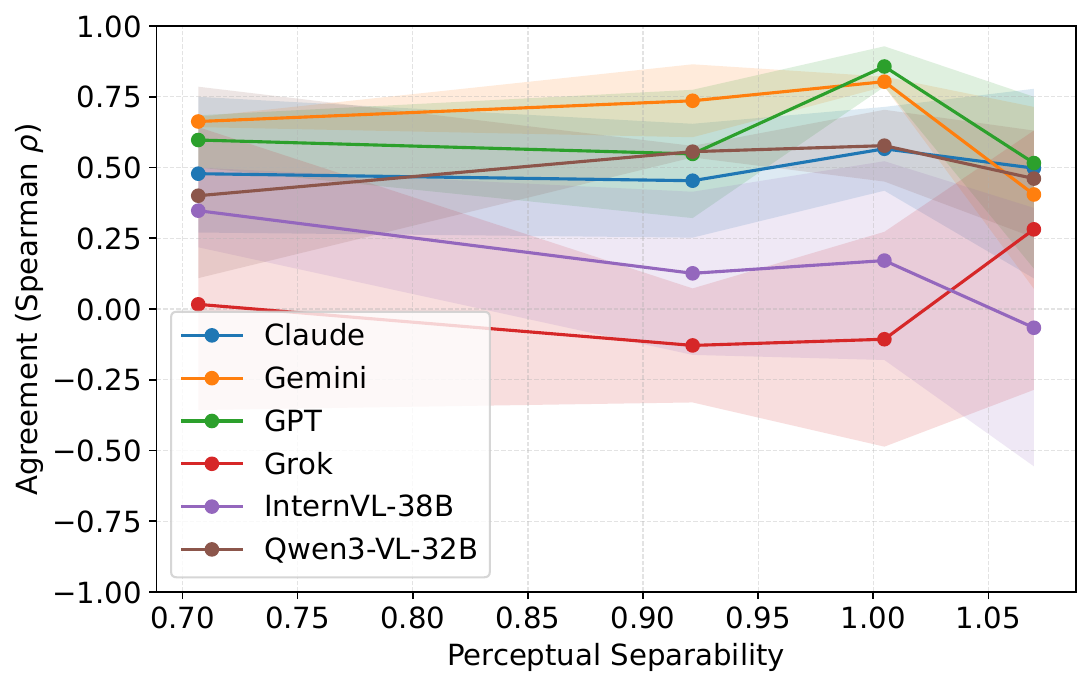}
    \caption*{(b) Contrast}
\end{minipage}%
\hfill
\begin{minipage}[t]{0.32\textwidth}
    \centering
    \includegraphics[width=\linewidth]{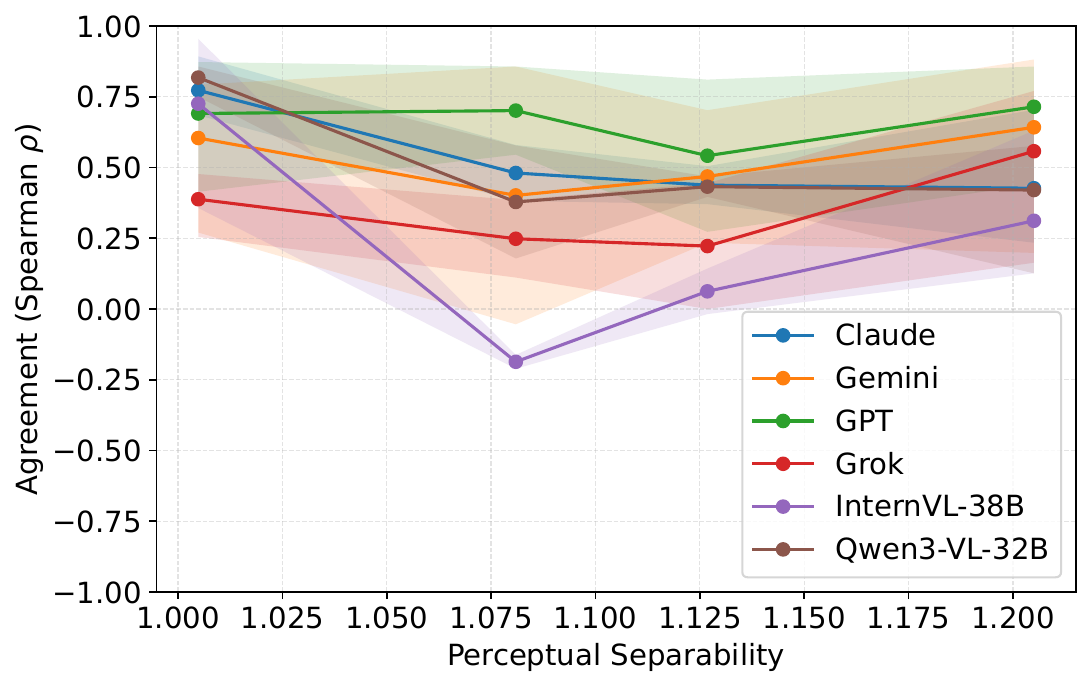}
    \caption*{(c) Overall preference}
\end{minipage}
\vspace{12pt}  

\caption{Mean scene agreement (Spearman $\rho$) as a function of perceptual separability for (a) Colorfulness, (b) Contrast, and (c) Overall Preference.}
\label{fig:difficulty_agreement}

\end{figure*}

\subsection{Attribute Weighting Regression}
To assess how overall preference relates to other attributes and human perception, overall preference was modeled as a linear combination of contrast and colorfulness using standardized linear regression: $z(\mathrm{Overall}) = \alpha, z(\mathrm{Contrast}) + \beta, z(\mathrm{Colorfulness})$, where $\alpha$ and $\beta$ represent their relative contributions, respectively. Each regression used matched (scene, image rendering model) observations to ensure that predictors correspond to the same rendered image rather than rank position.

Fig.~\ref{fig:attribute_weights} visualizes the resulting coefficients for humans and VLMs. Human judgments place greater weight on colorfulness than contrast, indicating that perceived color richness plays a stronger role than contrast when forming overall preference ratings.

Most of the VLMs exhibit similar attribute reliance as psychophysical data. Qwen, Gemini, and Intern show strong weighting toward colorfulness with relatively smaller contrast contributions, positioning them close to the human weighting pattern along the colorfulness axis. Claude also follows this trend but with slightly more balanced contributions from both attributes. In contrast, GPT assigns substantially higher weight to contrast while maintaining moderate sensitivity to colorfulness, indicating a more balanced but contrast-leaning strategy. Grok shows the most distinct behavior, relying heavily on contrast while assigning minimal weight to colorfulness.

\subsection{Perceptual Separability Analysis}

To analyze how VLM--human agreement varies with scene perceptual separability, a scene-level separability score was computed from the psychophysical data. Scene-wise rendering separability is defined as the standard deviation of psychophysical \(z\)-scores across rendering models for each scene, with higher values indicating greater perceptual discriminability.

For each scene \(s\), let \(\mathbf{h}_s\) and \(\mathbf{m}_s\) denote the psychophysical and model \(z\)-score vectors across rendering models, respectively. Agreement is quantified using Spearman's rank correlation \(a(s)=\rho(\mathbf{h}_s,\mathbf{m}_s)\), and its relationship with the separability score \(d(s)\) is analyzed across scenes. Uncertainty is estimated via bootstrap resampling over scenes, with 95\% confidence intervals obtained from the 2.5th and 97.5th percentiles.

Fig.~\ref{fig:difficulty_agreement} shows VLM--human agreement as a function of perceptual separability. For Colorfulness, agreement is generally high and stable for Claude, GPT, and Qwen, while Gemini improves as separability increases and Grok remains comparatively weak. For Contrast, agreement generally increases with separability for several models, particularly GPT and Gemini, whereas Grok and Intern show weaker and less stable agreement. For Overall Preference, agreement varies more substantially across models and separability levels; GPT remains relatively strong, while Intern exhibits the greatest variation. Overall, the results suggest that VLM--human agreement is influenced by perceptual separability, although the relationship is model- and scale-dependent.

\begin{figure}[!t]
  \centering
  \includegraphics[width=\columnwidth]{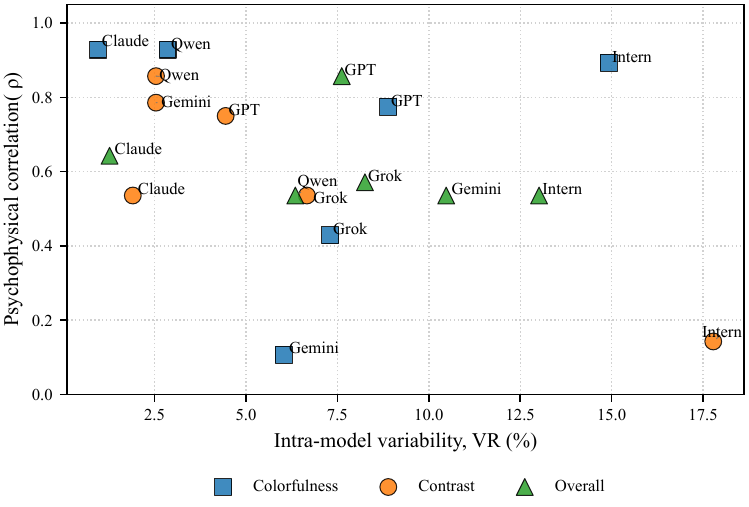}
  \caption{Relationship between intra-model variability (VR\%) and psychophysical correlation (Spearman $\rho$). }
  \label{fig:consistency_validity}
\end{figure}

\section{Perceptual Agreement vs Consistency}
Figure~\ref{fig:consistency_validity} examines the relationship between intra-model variability, measured as within-model disagreement (VR\%), and psychophysical rank correlation (Spearman $\rho$) across all models and perceptual scales. Each point represents a model--scale pair, with marker shapes indicating contrast, colorfulness, and overall preference.

A key observation is that low intra-model variability does not consistently correspond to higher agreement with human perception. While some models exhibit both high repeatability and strong psychophysical agreement (e.g., Claude and Qwen for colorfulness), this relationship is not systematic across scales. For instance, Claude achieves very low variability ($\approx 1$--$3$\% VR) and high correlation for colorfulness ($\rho \approx 0.93$), yet shows substantially lower agreement for contrast ($\rho \approx 0.54$) despite similarly high internal consistency. This indicates that self-consistency alone is not a reliable indicator of perceptual validity.

At the same time, models with higher variability do not necessarily perform poorly. GPT exhibits moderate variability ($\approx 7$--$9$\% VR) while achieving strong agreement with human rankings for overall preference ($\rho \approx 0.85$). In contrast, Intern shows high variability for contrast ($\approx 18$\% VR) together with very low correlation ($\rho \approx 0.14$), illustrating that high variability can coincide with poor perceptual alignment, but does not fully explain it.

Attribute-dependent effects are also evident. Colorfulness generally exhibits stronger agreement at lower variability, whereas contrast and overall preference show greater dispersion. This suggests that the relationship between consistency and perceptual validity is mediated by the perceptual scale rather than being an intrinsic property of the model.

This distinction between repeatability and perceptual validity is important when considering VLMs as surrogate observers, since stable model responses do not necessarily imply human-aligned judgments.

\section{Conclusion}
\label{sec:conclusion}

In this study, six VLMs were assessed against psychophysical image quality data using three image quality scales, i.e., contrast, colorfulness, and overall preference. The results show attribute-dependent perceptual agreement with direct implications for computational image quality assessment. The assessments vary substantially across both models and perceptual scales: GPT shows the most balanced cross-attribute behaviour and the strongest overall-preference agreement ($\rho = 0.86$), whereas Claude and Qwen show the highest colorfulness alignment ($\rho = 0.93$ each), and Qwen also achieves strong contrast agreement ($\rho = 0.86$). However, no current VLM consistently captures the full range of low-level and higher-level perceptual dimensions required for robust IQA.

These results indicate that VLMs can support rapid screening and exploratory evaluation in perceptual IQA workflows; however, current VLMs are not yet capable of substituting psychophysical experiments for rigorous perceptual evaluation. 

Given the limited set of 10 source scenes, future work should validate these findings on larger and more diverse datasets to assess content dependency and generalizability across tone reproduction, chroma variation, local contrast, and naturalness.

\section{Acknowledgment}
This publication has emanated from research jointly funded by Taighde Éireann -- Research Ireland under Grant Number 13/RC/2094\_2, and co-funded by the European Union under the Systems, Methods, Context (SyMeCo) programme Grant Agreement Number 101081459. Views and opinions expressed are however, those of the author(s) only and do not necessarily reflect those of the European Union or the European Research Executive Agency. Neither the European Union nor the granting authority can be held responsible for them.
{\small

}


{\small
\bibliographystyle{unsrt}
\bibliography{main}

@String(ICLR = {Int. Conf. Learn. Represent.})

@String(ICLR = {ICLR})

@article{mccamy1976color,
title={A color-rendition chart},
author={McCamy, Calvin S and Marcus, Harold and Davidson, James G and others},
journal={J. App. Photog. Eng},
volume={2},
number={3},
pages={95--99},
year={1976}
}

@book{ishihara1985ishihara,
title={Ishihara's test for colour-blindness},
author={Ishihara, Shinobu},
year={1985},
publisher={Kanehara Shuppan Company}
}

@article{fang2014no,
title={No-reference quality assessment of contrast-distorted images based on natural scene statistics},
author={Fang, Yuming and Ma, Kede and Wang, Zhou and Lin, Weisi and Fang, Zhijun and Zhai, Guangtao},
journal={IEEE Signal Processing Letters},
volume={22},
number={7},
pages={838--842},
year={2014},
publisher={IEEE}
}

@article{ghadiyaram2017perceptual,
title={Perceptual quality prediction on authentically distorted images using a bag of features approach},
author={Ghadiyaram, Deepti and Bovik, Alan C},
journal={Journal of vision},
volume={17},
number={1},
pages={32--32},
year={2017},
publisher={The Association for Research in Vision and Ophthalmology}
}

@article{xu2016blind,
title={Blind image quality assessment based on high order statistics aggregation},
author={Xu, Jingtao and Ye, Peng and Li, Qiaohong and Du, Haiqing and Liu, Yong and Doermann, David},
journal={IEEE Transactions on Image Processing},
volume={25},
number={9},
pages={4444--4457},
year={2016},
publisher={IEEE}
}

@article{xue2014blind,
title={Blind image quality assessment using joint statistics of gradient magnitude and Laplacian features},
author={Xue, Wufeng and Mou, Xuanqin and Zhang, Lei and Bovik, Alan C and Feng, Xiangchu},
journal={IEEE Transactions on Image Processing},
volume={23},
number={11},
pages={4850--4862},
year={2014},
publisher={IEEE}
}

@article{mittal2012no,
title={No-reference image quality assessment in the spatial domain},
author={Mittal, Anish and Moorthy, Anush Krishna and Bovik, Alan Conrad},
journal={IEEE Transactions on image processing},
volume={21},
number={12},
pages={4695--4708},
year={2012},
publisher={IEEE}
}

@article{mehmood2022reference,
author={Mehmood, Imran and Liu, Xiaoxuan and Khan, Muhammad Usman and Luo, Ming Ronnier},
title={Method for developing and using high quality reference images to evaluate tone mapping operators},
journal={Journal of the Optical Society of America A},
volume={39},
number={6},
pages={B11--B20},
year={2022}
}

@misc{Anthropic_ClaudeOpus46_2026,
author={{Anthropic}},
title={{Claude Opus 4.6 (2026 version) [Large Language Model]}},
year={2026},
note={Accessed Feb 13, 2026}
}

@misc{Google_Gemini31Pro_2026,
author={{Google}},
title={{Gemini 3.1 Pro (Gemini) [Large Language Model]}},
year={2026},
howpublished={\url{https://deepmind.google/technologies/gemini/}},
note={Accessed Feb 14, 2026}
}

@misc{OpenAI_GPT52_2026,
author={{OpenAI}},
title={{GPT-5.2 (GPT) [Large Language Model]}},
year={2025},
note={Accessed Feb 14, 2026}
}

@misc{xAI_Grok41_2026,
author={{xAI}},
title={{Grok-4.1 (Grok) [Large Language Model]}},
year={2026},
note={Accessed Feb 14, 2026}
}

@misc{OpenGVLab_InternVL35_38B_2026,
author={{OpenGVLab}},
title={{InternVL-3.5-38B ((Jan 2026 version)) [Vision-Language Model]}},
year={2025},
note={Accessed Feb 20, 2026}
}

@misc{QwenVL3_2026,
author={{Alibaba}},
title={{QwenVL3 (Jan 2026 version) [Vision Language Model]}},
year={2026},
note={Accessed Feb 24, 2026}
}

@article{berns1996,
author={Berns, Roy S.},
title={Methods for characterizing CRT displays},
journal={Displays},
volume={16},
number={4},
pages={173--182},
year={1996}
}

@techreport{itu2020,
author={{International Telecommunication Union}},
title={Methodologies for the subjective assessment of the quality of television images},
year={2020},
institution={International Telecommunication Union}
}

@article{khan2020,
author={Khan, I. R. and Aziz, W. and Shim, S.-O.},
title={Tone-mapping using perceptual-quantizer and image histogram},
journal={IEEE Access},
volume={8},
pages={31350--31358},
year={2020}
}

@article{hui2018,
author={Li, Hui and Jia, Xixi and Zhang, Lei},
title={Clustering based content and color adaptive tone mapping},
journal={Computer Vision and Image Understanding},
volume={168},
pages={37--49},
year={2018}
}

@inproceedings{liang2018,
author={Liang, Z. and Xu, J. and Zhang, D. and Cao, Z. and Zhang, L.},
title={A hybrid L1-L0 layer decomposition model for tone mapping},
booktitle={Proceedings of the IEEE Conference on Computer Vision and Pattern Recognition},
pages={4758--4766},
year={2018}
}

@article{mehmood2023perceptual,
author={Mehmood, Imran and Shi, Xinye and Khan, Muhammad Usman and Luo, Ming Ronnier},
title={Perceptual tone mapping model for high dynamic range imaging},
journal={IEEE Access},
volume={11},
pages={110272--110288},
year={2023}
}

@article{meylan2006,
author={Meylan, Laurence and S{"u}sstrunk, Sabine},
title={High dynamic range image rendering with a retinex-based adaptive filter},
journal={IEEE Transactions on Image Processing},
volume={15},
number={9},
pages={2820--2830},
year={2006}
}

@inproceedings{reinhard2002,
author={Reinhard, Erik and Stark, Michael and Shirley, Peter and Ferwerda, James},
title={Photographic tone reproduction for digital images},
booktitle={Proceedings of the 29th Annual Conference on Computer Graphics and Interactive Techniques},
pages={267--276},
year={2002}
}

@inproceedings{schlick1995,
author={Schlick, Christophe},
title={Quantization techniques for visualization of high dynamic range pictures},
booktitle={Photorealistic Rendering Techniques},
pages={7--20},
publisher={Springer},
year={1995}
}

@article{thurstone1927,
author={Thurstone, L. L.},
title={A law of comparative judgment},
journal={Psychological Review},
volume={34},
number={4},
pages={273--286},
year={1927}
}

@article{chow2016review,
title={Review of medical image quality assessment},
author={Chow, L S and Paramesran, R},
journal={Biomedical Signal Processing and Control},
year={2016}
}

@inproceedings{wu2024qbench,
title={Q-Bench: Benchmark for low-level vision},
author={Wu, Haoning and others},
booktitle={ICLR},
year={2024}
}

@article{kil2024mllm,
title={MLLM comparative reasoning for vision},
author={Kil, J and others},
journal={arXiv},
year={2024}
}

@inproceedings{kim2026vlmsubtlebench,
title={VLM-SubtleBench},
author={Kim, M and others},
booktitle={ICLR},
year={2026}
}

@article{khan2026no,
title={No-Reference Generic Image Quality Assessment via Adaptive Fusion of Naturalness-Enhanced Colour and Appearance-Driven Spatial Metrics},
author={Khan, Muhammad Usman and Mehmood, Imran and Luo, Ming Ronnier},
journal={IEEE Access},
year={2026},
publisher={IEEE}
}

@inproceedings{zhang2018unreasonable,
title={The unreasonable effectiveness of deep features as a perceptual metric},
author={Zhang, Richard and Isola, Phillip and Efros, Alexei A and Shechtman, Eli and Wang, Oliver},
booktitle={2018 IEEE/CVF conference on computer vision and pattern recognition},
pages={586--595},
year={2018},
organization={IEEE}
}

@article{preiss2014color,
title={Color-image quality assessment: From prediction to optimization},
author={Preiss, Jens and Fernandes, Felipe and Urban, Philipp},
journal={IEEE Transactions on Image Processing},
volume={23},
number={3},
pages={1366--1378},
year={2014},
publisher={IEEE}
}

@article{mehmood2024generic,
  title={Generic color correction for tone mapping operators in high dynamic range imaging},
  author={Mehmood, Imran and Khan, Muhammad Usman and Luo, Ming Ronnier},
  journal={Optics Express},
  volume={32},
  number={16},
  pages={27849--27866},
  year={2024},
  publisher={Optica Publishing Group}
}

@String(ICLR  = {ICLR})
}

\end{document}